\documentclass[10pt,twocolumn,letterpaper]{article}

\usepackage{iccv}
\usepackage{times}
\usepackage{epsfig}
\usepackage{graphicx}
\usepackage{amsmath}
\usepackage{amssymb}
\usepackage{setspace}
\usepackage{booktabs}
\usepackage{multirow}
\usepackage[accsupp]{axessibility}

\usepackage[breaklinks=true,bookmarks=false]{hyperref}

\iccvfinalcopy 

\def\iccvPaperID{2902} 

\ificcvfinal\fi

\begin{document}

\title{Cross-category Video Highlight Detection via Set-based Learning}

\author{
	Minghao Xu\textsuperscript{\rm 1} \quad
	Hang Wang\textsuperscript{\rm 1} \quad
	Bingbing Ni\textsuperscript{\rm 1}\footnotemark[1] \quad
	Riheng Zhu\textsuperscript{\rm 2} \quad
	Zhenbang Sun\textsuperscript{\rm 2} \quad
	Changhu Wang\textsuperscript{\rm 2}\\
	\textsuperscript{\rm 1}Shanghai Jiao Tong University, Shanghai 200240, China \quad
	\textsuperscript{\rm 2}ByteDance AI Lab \\
	\{xuminghao118, wang--hang, nibingbing\}@sjtu.edu.cn \\
	\{zhuriheng, sunzhenbang, wangchanghu\}@bytedance.com
}

\maketitle
\ificcvfinal\thispagestyle{empty}\fi


\begin{abstract}

Autonomous highlight detection is crucial for enhancing the efficiency of video browsing on social media platforms. To attain this goal in a data-driven way, one may often face the situation where highlight annotations are not available on the target video category used in practice, while the supervision on another video category (named as source video category) is achievable. In such a situation, one can derive an effective highlight detector on target video category by transferring the highlight knowledge acquired from source video category to the target one. We call this problem cross-category video highlight detection, which has been rarely studied in previous works. For tackling such practical problem, we propose a \textbf{D}ual-\textbf{L}earner-based \textbf{V}ideo \textbf{H}ighlight \textbf{D}etection (DL-VHD) framework. Under this framework, we first design a \textbf{S}et-based \textbf{L}earning module (SL-module) to improve the conventional pair-based learning by assessing the highlight extent of a video segment under a broader context. Based on such learning manner, we introduce two different learners to acquire the basic distinction of target category videos and the characteristics of highlight moments on source video category, respectively. These two types of highlight knowledge are further consolidated via knowledge distillation. Extensive experiments on three benchmark datasets demonstrate the superiority of the proposed SL-module, and the DL-VHD method outperforms five typical Unsupervised Domain Adaptation (UDA) algorithms on various cross-category highlight detection tasks. Our code is available at \url{https://github.com/ChrisAllenMing/Cross_Category_Video_Highlight}.

\end{abstract}


\vspace{-3mm}
\section{Introduction} \label{sec1}
\vspace{-0.5mm}

In current days, people show growing interests in sharing the videos recording their daily life on the social media platforms like \emph{YouTube} and \emph{Instagram}\footnotetext[0]{*Corresponding author: Bingbing Ni.}. Among all these videos, the well-edited ones that summarize the highlights of specific events are apparently more attractive to the audience. However, in most cases, the original video of a real-world event contains many contents unrelated to its gist, and it is an onerous and time-consuming task to pick out the highlight parts of the video manually. Therefore, in order to enhance the efficiency of video content refinement, it is desirable to develop a machine learning model for autonomous video highlight detection.


\begin{figure}[t]
    \centering
    \includegraphics[width=0.47\textwidth]{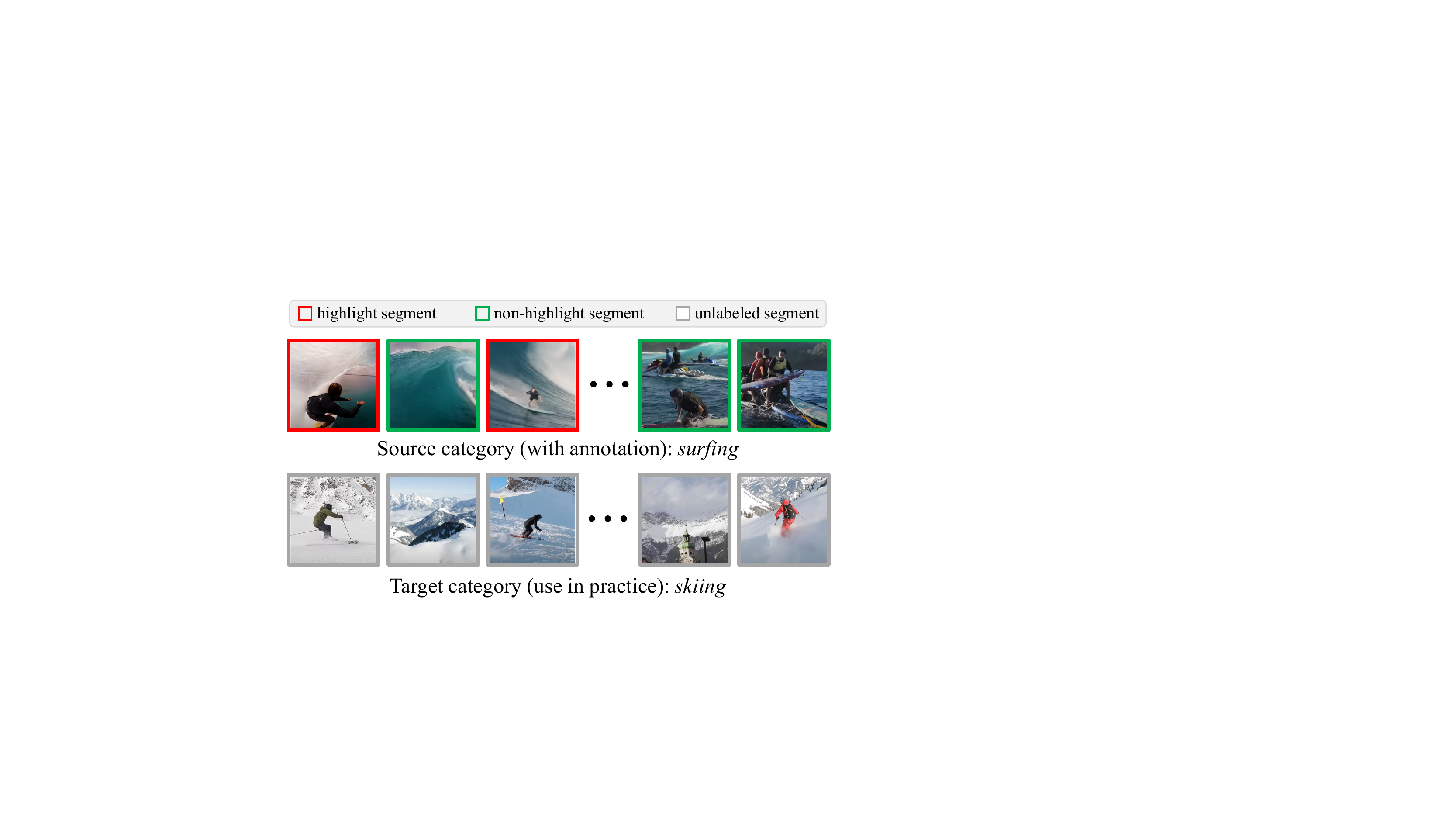}
    \caption{The situation in which the target video category used in practice lacks supervision, while another video category, \emph{i.e.} the source one, possesses annotation.} 
    \label{fig_motivation}
\vspace{-3mm}
\end{figure}


To endow a model with the capability of identifying the highlight segments within a video, existing works have explored various ways of supervision, including the explicit highlight annotations~\cite{video2gif,deep_rank,3d_deep_rank}, the frequent occurrence of specific video segments~\cite{category-specific,robust_autoencoder,summarize_web_video}, the duration of a video~\cite{learn_from_duration}, \emph{etc.} These approaches generally focused on training a highlight detector for a specific video category (\emph{e.g.} surfing, skiing, parkour, \emph{etc.}), while the transferability of a highlight detection model across different video categories has been less studied in previous works.

As a matter of fact, in practical applications, one can face the situation where supervisory signal is lacked on the target video category intended to be used in practice, while the supervision on another video category is available, just as shown in Fig.~\ref{fig_motivation}. Under such situation, we consider the problem of \emph{Cross-category Video Highlight Detection}. The setting of this problem is analogous to that of Unsupervised Domain Adaptation (UDA)~\cite{survey} in which one seeks to adapt the knowledge learned from the labeled source domain (the source video category with supervision) to the unlabeled target domain (the unsupervised target video category). 

In addition, for optimizing the highlight detector, most of existing methods~\cite{video2gif,deep_rank,3d_deep_rank,category-specific,learn_from_duration,mini-net} followed the philosophy of pair-based learning, \emph{i.e.} comparing a positive sample (\emph{e.g.} a highlight video segment or a segment bag containing highlights) with a negative one, and, after training, the former is expected to rank higher than the latter. Nevertheless, such learning manner might not fully exploit the contextual information spanning among different video segments. For example, in a soccer match, the moment of a player's dribbling the ball is more attractive than the one of the players' entering the pitch, and both of them are less exciting than the moment of a goal. These relationships can hardly be captured by a single segment pair, which makes the highlight prediction of a pair-learning-based model potentially imprecise in the span of a whole video.

Motivated by the facts above, in this work, we propose a \textbf{D}ual-\textbf{L}earner-based \textbf{V}ideo \textbf{H}ighlight \textbf{D}etection (\emph{DL-VHD}) framework to address the Cross-category Video Highlight Detection problem. Under this framework, we first devise a \textbf{S}et-based \textbf{L}earning module (SL-module) to improve the conventional pair-based learning manner for highlight detection. In a nutshell, this module learns to regress the highlight score distribution over a set of segments from the same video, in which a Transformer encoder~\cite{transformer} is employed to model the interrelationship among various video segments. Based on this learning mechanism, we further introduce two different learners to capture two types of knowledge about highlight moments. In specific, the \emph{coarse-grained learner} gains the basic concepts about what distinguishes the videos of target category from other ones, and the \emph{fine-grained learner} acquires the precise highlight notions on source videos. These two kinds of knowledge are further integrated by distilling each of them into the other learner, and such integrated knowledge forms the more complete concepts about the highlight moments on target video category. In practice, the SL-module can be individually applied to derive an effective highlight detector when the segment-level annotation is available on the target video category, while, when such annotation is unobtainable, we can resort to the DL-VHD method for highlight knowledge transfer. 

Our contributions can be summarized as follows:
\begin{itemize}
    \item To the best of our knowledge, this work is the first attempt at cross-category video highlight detection, in which we utilize a dual-learner-based scheme to transfer the concepts about highlight moments across different video categories. 
    \item We propose a novel set-based learning mechanism which is able to identify whether a video segment is highlight or not under a broader context. 
    \item Under the category-specific setting, we verify the superior performance of the SL-module over previous methods. For cross-category highlight detection, the DL-VHD model substantially surpasses existing UDA algorithms and performs comparably with the supervised model trained on target video category.
\end{itemize}


\section{Related Work} \label{sec2}

\textbf{Video Highlight Detection.} This task aims at assigning each video segment a score of its worthiness as highlight. In recent years, the videos studied for this task extend from sport videos~\cite{TV_baseball,audio-visual_marker,cricket} to general videos from social media~\cite{youtube_highlights} or first-person camera shooting~\cite{deep_rank}. According to the manner of supervision, the existing works on this topic can be generally divided into two classes. For the supervised methods~\cite{video2gif,deep_rank,3d_deep_rank,youtube_highlights}, the highlight annotations of all segments in a video are given. For the weakly-supervised approaches~\cite{category-specific,robust_autoencoder,summarize_web_video,learn_from_duration,mini-net}, various weak supervisory signals have been exploited to define highlights, including the frequent occurrence of specific segments within a video category ~\cite{category-specific,robust_autoencoder,summarize_web_video}, the duration of a video~\cite{learn_from_duration} and the information from segment bags~\cite{mini-net}. For model optimization, most of these methods~\cite{video2gif,deep_rank,3d_deep_rank,youtube_highlights,learn_from_duration,mini-net} followed the philosophy of pair-based learning, \emph{i.e.} comparing between a positive sample and a negative one.


\emph{Improvements over existing methods.} 
In this work, we novelly explore the cross-category video highlight detection problem through learning two types of knowledge about highlight moments and integrating them on target video category. In addition, a set-based learning mechanism is proposed to improve the pair-based learning by performing highlight prediction on a set of video segments, such that the highlight extent of each segment can be judged more precisely with rich contextual information.


\textbf{Unsupervised Domain Adaptation (UDA).} UDA focuses on generalizing a model learned from the labeled source domain to another unlabeled target domain. To pursue this goal, a commonly used strategy is to minimize a specific metric for measuring domain shift~\cite{da_theory,survey}, \emph{e.g.} Maximum Mean Discrepancy (MMD)~\cite{two-sample_test,domain_confusion}, Multi-Kernel MMD~\cite{dan}, Weighted MMD~\cite{w-mmd}, Wasserstein Distance~\cite{w-distance,sliced_w-distance} and the difference of feature covariance~\cite{deepcoral} or feature norm~\cite{larger_norm}. On another line of research, adversarial learning is employed to facilitate domain-invariance on either pixel level~\cite{pixel-level,gta,cycada} or feature level~\cite{revgrad,adda,cdan,domain_mixup}. In order to introduce the discriminative information on target domain, recent works~\cite{semantic,progressive,minimax_entropy,gpa,ltc-msda,graph-msda} utilized the pseudo labels of target samples for category-level domain alignment. This work explores cross-category video highlight detection, a similar problem as UDA, in which one intends to transfer the highlight knowledge acquired from the source video category to the target one. 




\section{Method} \label{sec3}

In the cross-category video highlight detection problem, a set of videos containing the highlight moments of source video category, \emph{i.e.} $D_{\mathcal{S}} = \{v^{\mathcal{S}}_k\}_{k=1}^{|D_{\mathcal{S}}|}$, are given, and each video $v \in D_{\mathcal{S}}$ is divided into $N_v$ segments $\{(s_i, y_i)\}_{i=1}^{N_v}$ with similar duration, where $y_i$ denotes the ground-truth highlight label for segment $s_i$. In addition, we have another set of videos including the highlight moments of target video category, \emph{i.e.} $D_{\mathcal{T}} = \{v^{\mathcal{T}}_k\}_{k=1}^{|D_{\mathcal{T}}|}$, while the segment-level highlight annotations of target category are not available on these videos. Under such condition, the main objective is to derive an effective highlight detector on target video category through fully exploiting the labeled source videos and the unlabeled target ones.


\subsection{Motivation and Overview} \label{sec3_1}

\textbf{Cross-category Video Highlight Detection.} In real-world applications, the segment-level highlight annotations may not be available for the target video category that the model is applied to, while one can obtain the supervision on another video category (named as source video category). Therefore, in such a situation, a natural question to ask is how to transfer the knowledge about highlight moments on source video category to the target one, \emph{i.e.}~performing \emph{cross-category video highlight detection}. A straightforward answer is to leverage the existing Unsupervised Domain Adaptation (UDA) techniques for feature distribution alignment between two distinct video categories. However, such distribution alignment is hard, if not ill-posed, for the highlight detection problem, since the highlight segments for the target category may be nuisance for the source one, and vice versa, which is experimentally illustrated in Sec.~\ref{sec4_3}.

To acquire the exact highlight concepts for target video category using the data from both categories, we propose a \textbf{D}ual-\textbf{L}earner-based \textbf{V}ideo \textbf{H}ighlight \textbf{D}etection (\emph{DL-VHD}) framework. Under this framework, the model learns two kinds of knowledge about highlight moments, \emph{i.e.} the distinction of target category videos with other ones and the characteristics of highlights on source category. These two types of knowledge are further merged to form the more complete highlight concepts about target video category.  


\textbf{Set-based Learning.} Previous works~\cite{video2gif,deep_rank,youtube_highlights,learn_from_duration,mini-net} commonly trained the highlight detection model by contrasting a highlight segment $s_{+}$ with a non-highlight segment $s_{-}$, which seeks to model the conditional distribution $p(y_{+}, y_{-} | s_{+}, s_{-})$. 
However, such pair-based learning may fail to discover the more complex highlight relations among more than two segments. For example, the excitement level of a soccer match differs from moment to moment, and the relative highlight extent of these moments cannot be sufficiently captured by pairs of video segments. 

Motivated by such limitation, we propose a \textbf{S}et-based \textbf{L}earning module (SL-module).
Its core idea is to train the model to predict the highlight score distribution over a set of video segments, and the prediction of a single segment is depended on all the other segments in the set, which models $p(y_{1}, y_{2}, \cdots, y_{N} | s_{1}, s_{2}, \cdots, s_{N})$ ($N$ denotes the set size). By including such contextual information spanning among different video segments, it is expected that the model can assign more accurate highlight score to each segment.


\textbf{Cross-category Video Highlight Detection via Set-based Learning.} In order to bridge the highlight patterns of two distinct video categories, it is essential to explore the interrelationship among the video segments within the same category and also across different categories. Such complex relational patterns can be better captured under the rich context provided by segment sets. Based on such motivation, in DL-VHD, we employ SL-module as the basic learning module to acquire more precise highlight knowledge.  


\begin{figure}[t]
    \centering
    \includegraphics[width=0.48\textwidth]{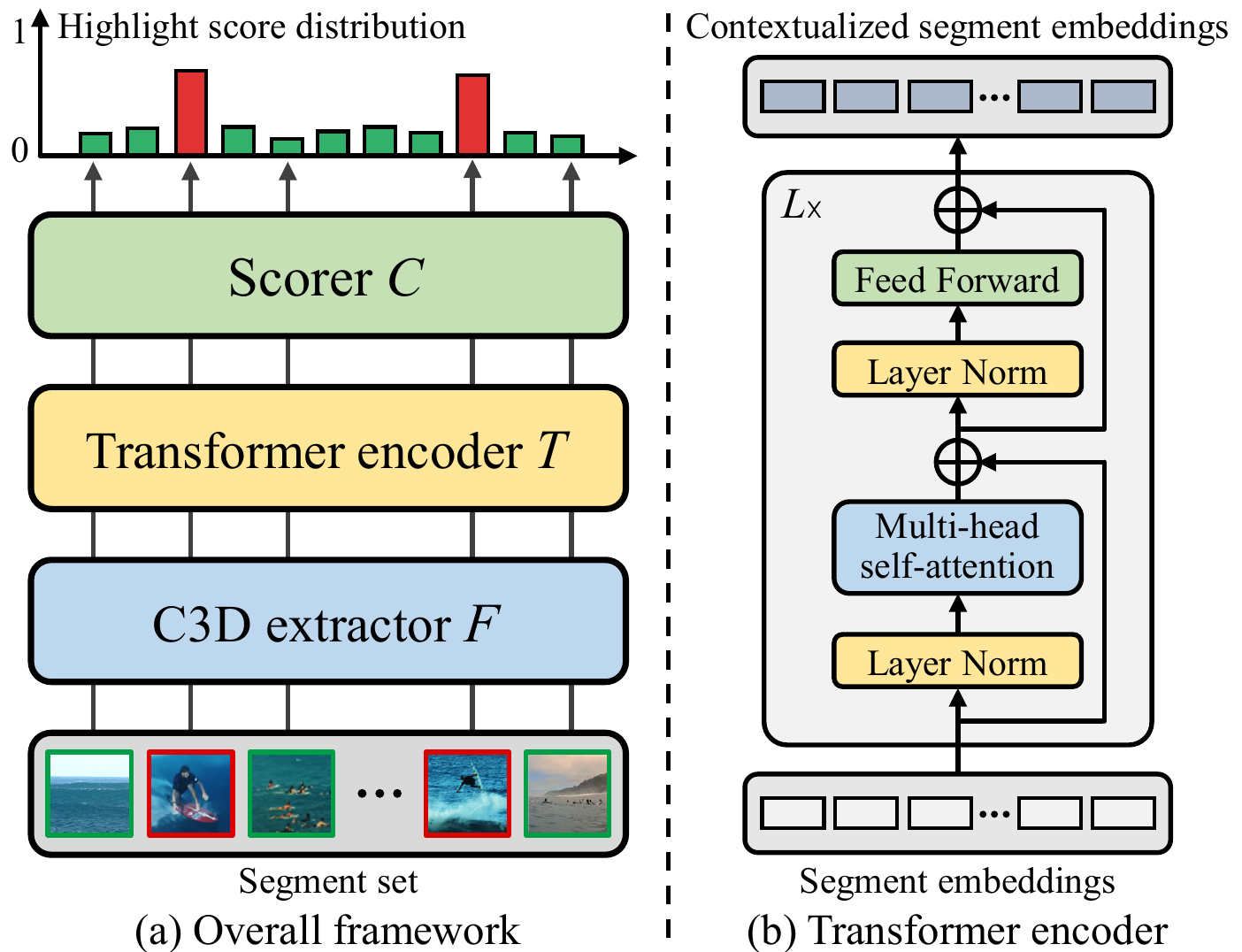}
    \caption{(a) The overall framework of SL-module. (b) The architecture of the Transformer encoder used in this module.} 
    \label{fig_set_learning}
\vspace{-3mm}
\end{figure}


\begin{figure*}[t]
    \centering
    \includegraphics[width=0.96\textwidth]{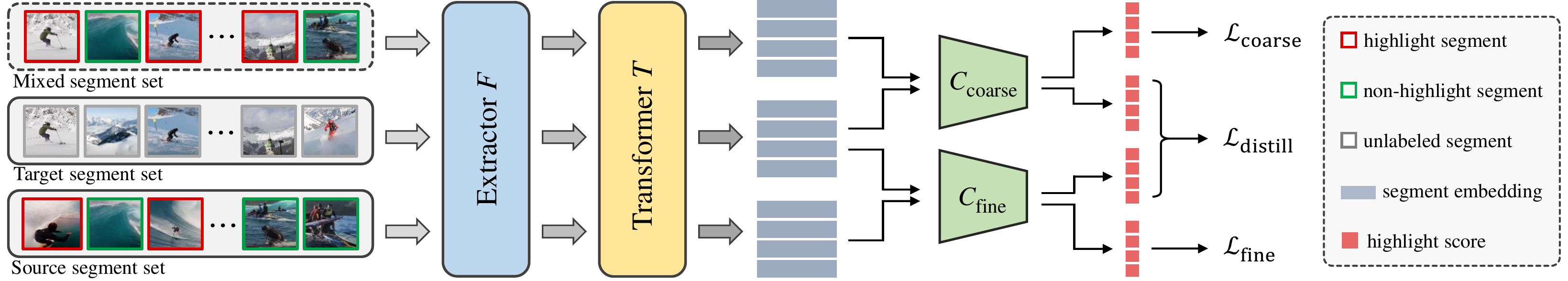}
    \caption{\textbf{Illustration of DL-VHD.} Three kinds of segment sets are constructed with a labeled source video and an unlabeled target video, and the segment embeddings are derived by a C3D extractor and a Transformer encoder. A coarse-grained and a fine-grained learner are supervised by the mixed and source segment set, respectively, and their knowledge is further consolidated by knowledge distillation.} 
    \label{fig_dual_learner}
\vspace{-3mm}
\end{figure*}


\subsection{Set-based Learning Module} \label{sec3_2}

The SL-module models the interdependency among the video segments in a set and predicts the highlight score of each segment under such set-determined context, as shown in Fig.~\ref{fig_set_learning}(a). Next, we introduce the detailed learning and inference schemes of this module. 


\textbf{Learning scheme.} In each learning step, a set of $N$ annotated segments randomly sampled from the same video, \emph{i.e.} $x = \{(s_j, y_j)\}_{j=1}^{N}$, is given, and a pre-trained C3D~\cite{c3d} model $F$ extracts the feature embedding of each segment, \emph{i.e.} $z = \{z_j\}_{j=1}^{N} = \{F(s_j)\}_{j=1}^{N}$ ($F$ is fixed in the learning phase). On these segment embeddings, a Transformer encoder~\cite{transformer} $T$ models the interrelationship among different segments and outputs the contextualized segment embeddings, \emph{i.e.} $\tilde{z} = \{\tilde{z}_j\}_{j=1}^{N} = T(z)$. The Transformer encoder used in our method basically follows the original design in \cite{transformer} which stacks $L$ layers of multi-head self-attention and feed forward network. In contrast, we remove the positional encoding module for the permutation-invariance of set learning, and, as suggested in \cite{pre-norm}, Layer Normalization (LN)~\cite{layer_norm} is applied before each self-attention and feed forward module. The architecture of the Transformer encoder is shown in Fig.~\ref{fig_set_learning}(b). We refer readers to the original literature~\cite{transformer} for more details. 

Upon the contextualized segment embeddings, a scoring model $C$ predicts the highlight score of each video segment, \emph{i.e.} $\hat{y} = \{\hat{y}_j\}_{j=1}^{N} = \{C(\tilde{z}_j)\}_{j=1}^{N}$. Now that the highlight prediction on a segment set is obtained, we define the learning objective. During the learning phase, the basic desiderata is to match the highlight score distribution on set $x$ predicted by the model with the ground-truth distribution. To attain this goal, we define the learning objective as follows:
\begin{equation} \label{eq1}
\min \limits_{T,C} \ \mathcal{L}_{\mathrm{pred}} , 
\end{equation}
\begin{equation} \label{eq2}
\mathcal{L}_{\mathrm{pred}} = D_{\mathrm{KL}}\Big( \sigma \big ( \{y_j\}_{j=1}^{N} \big), \sigma \big( \{\hat{y}_j\}_{j=1}^{N} \big) \Big) ,
\end{equation}
where $\sigma(\cdot)$ denotes the softmax function producing the predicted and ground-truth highlight score distributions, and $D_{\mathrm{KL}}(\cdot, \cdot)$ stands for the Kullback–Leibler divergence.


\textbf{Inference scheme.} To infer the highlight score of a segment $s$ from a test video, we first construct a segment set containing $s$ and other $N-1$ context segments adjacent to $s$ in the video, denoted as $x = \{s_j\}_{j=1}^{N}$. Half of these context segments are right before $s$ in the video, and the other half are right after. Segment duplication is conducted if the segments around $s$ cannot fill the set.
Here, we do not use the set composed of random segments as in the learning phase for the sake of suppressing the variance of prediction. We then infer the highlight score of all segments in the set, \emph{i.e.} $\hat{y} = \{\hat{y}_j\}_{j=1}^{N}$, by feeding them into the C3D feature extractor, the Transformer encoder and the scoring model successively. Finally, we pick out $\hat{y}_{\mathrm{id}(s)}$ ($\mathrm{id}(s)$ stands for the index of $s$ in $x$) as the highlight score of the video segment to be evaluated. 


\subsection{Dual-Learner-based Video Highlight Detection} \label{sec3_3}

On the basis of SL-module, we now explore the cross-category video highlight detection problem. Its main objective is to derive an effective highlight detector on target video category by fully exploiting the labeled source videos $D_{\mathcal{S}}$ and the unlabeled target ones $D_{\mathcal{T}}$. To pursue such goal, we seek to capture the highlight concepts about target video category from two aspects. On one hand, there exists some obvious features that distinguish the target category videos from the ones of other topics, \emph{e.g.} the surfboard in surfing videos, the ski pole in skiing videos, \emph{etc.} The perception of such features endows a model with the basic capability of picking out the segments of the target category from a video mixing different contents. On the other hand, there are some common characteristics of highlight moments sharing between distinct video categories. For instance, the moments with a standing person moving on some surface of the scene can be the highlights of both surfing and skiing videos. Such generic knowledge can be employed to identify the highlight moments for the target video category. However, neither of these two types of concepts alone can sufficiently define the highlights on the target category, which calls for a scheme that integrates different knowledge. 

Following the above intuitions, we design a dual-learner-based framework, in which two kinds of highlight knowledge are learned by a \emph{coarse-grained learner} and a \emph{fine-grained learner} respectively, and they are further integrated by a knowledge distillation~\cite{knowledge_distillation} scheme. The graphical illustration of this framework is shown in Fig.~\ref{fig_dual_learner}. We state the detailed learning and inference schemes as follows.


\textbf{Learning scheme.} In each learning step, we use a set of labeled segments randomly sampled from a video in $D_{\mathcal{S}}$, denoted as $x_{\mathcal{S}} = \{(s^{\mathcal{S}}_j, y^{\mathcal{S}}_j)\}_{j=1}^{N}$, and a set of unlabeled segments randomly sampled from a video in $D_{\mathcal{T}}$, denoted as $x_{\mathcal{T}} = \{s^{\mathcal{T}}_j\}_{j=1}^{N}$. Based on these two sets, we further construct a mixed set with the segments from two video categories, denoted as $x_{\mathcal{M}} = \{(s^{\mathcal{M}}_j, y^{\mathcal{M}}_j)\}_{j=1}^{N}$ ($y^{\mathcal{M}}_j$ equals to $1$ if $s^{\mathcal{M}}_j$ is a target category segment and is $0$ otherwise), in which half of the segments are randomly sampled from $x_{\mathcal{S}}$, and the other half are from $x_{\mathcal{T}}$. Using the C3D feature extractor and Transformer encoder, we respectively derive the contextualized segment embeddings for these three sets, \emph{i.e.} $\tilde{z}_{\mathcal{S}} = \{\tilde{z}^{\mathcal{S}}_j\}_{j=1}^{N}$, $\tilde{z}_{\mathcal{T}} = \{\tilde{z}^{\mathcal{T}}_j\}_{j=1}^{N}$ and $\tilde{z}_{\mathcal{M}} = \{\tilde{z}^{\mathcal{M}}_j\}_{j=1}^{N}$. 

Upon these segment embeddings, on one hand, we introduce a \emph{coarse-grained learner} $C_{\mathrm{coarse}}$ to learn the basic distinction of target video segments with the source ones. This is achieved by matching the highlight prediction of $C_{\mathrm{coarse}}$ on mixed set, \emph{i.e.} $\hat{y}_{\mathcal{M}} = \{\hat{y}^{\mathcal{M}}_j\}_{j=1}^{N} = \{C_{\mathrm{coarse}}(\tilde{z}^{\mathcal{M}}_j)\}_{j=1}^{N}$, with the ground-truth highlight distribution on that set, which defines the coarse-grained highlight prediction loss:
\begin{equation} \label{eq3}
\mathcal{L}_{\mathrm{coarse}} = D_{\mathrm{KL}}\Big( \sigma \big ( \{y^{\mathcal{M}}_j\}_{j=1}^{N} \big), \sigma \big( \{\hat{y}^{\mathcal{M}}_j\}_{j=1}^{N} \big) \Big) .
\end{equation}

On the other hand, a \emph{fine-grained learner} $C_{\mathrm{fine}}$ is introduced to acquire the knowledge about highlight moments on the source video category. This is attained through the supervised learning on set $x_{\mathcal{S}}$, in which the prediction of $C_{\mathrm{fine}}$, \emph{i.e.} $\hat{y}_{\mathcal{S}} = \{\hat{y}^{\mathcal{S}}_j\}_{j=1}^{N} = \{C_{\mathrm{fine}}(\tilde{z}^{\mathcal{S}}_j)\}_{j=1}^{N}$, is aligned with the ground-truth highlight score distribution:
\begin{equation} \label{eq4}
\mathcal{L}_{\mathrm{fine}} = D_{\mathrm{KL}}\Big( \sigma \big ( \{y^{\mathcal{S}}_j\}_{j=1}^{N} \big), \sigma \big( \{\hat{y}^{\mathcal{S}}_j\}_{j=1}^{N} \big) \Big) .
\end{equation}

Now that two types of knowledge about highlight moments are acquired by two different learners, we aim to integrate them on the target video category. Inspired by the idea of knowledge distillation~\cite{knowledge_distillation}, we would like to distill the knowledge of each learner into the other learner without impairing its original knowledge. Specifically, the coarse-grained and fine-grained learner are both utilized to predict the highlight scores of the segments in set $x_{\mathcal{T}}$, which gives out $\hat{y}_{\mathcal{T},\mathrm{coarse}} = \{\hat{y}^{\mathcal{T},\mathrm{coarse}}_j\}_{j=1}^{N} = \{C_{\mathrm{coarse}}(\tilde{z}^{\mathcal{T}}_j)\}_{j=1}^{N}$ and $\hat{y}_{\mathcal{T},\mathrm{fine}} = \{\hat{y}^{\mathcal{T},\mathrm{fine}}_j\}_{j=1}^{N} = \{C_{\mathrm{fine}}(\tilde{z}^{\mathcal{T}}_j)\}_{j=1}^{N}$. We then generate the prediction reflecting both kinds of highlight knowledge by averaging $\hat{y}_{\mathcal{T},\mathrm{coarse}}$ and $\hat{y}_{\mathcal{T},\mathrm{fine}}$, which produces $\hat{y}_{\mathcal{T},\mathrm{avg}} = \{\hat{y}^{\mathcal{T},\mathrm{avg}}_j\}_{j=1}^{N} = \{(\hat{y}^{\mathcal{T},\mathrm{coarse}}_j + \hat{y}^{\mathcal{T},\mathrm{fine}}_j) / 2\}_{j=1}^{N}$. In order to perform knowledge distillation between two learners, we constrain the individual prediction from either the coarse-grained or fine-grained learner to approach the average prediction, which defines the distillation loss as below:
\begin{equation} \label{eq5}
\begin{split}
\mathcal{L}_{\mathrm{distill}} = & \frac{1}{2} \bigg( D_{\mathrm{KL}} \Big( \sigma \big ( \{\hat{y}^{\mathcal{T},\mathrm{avg}}_j\}_{j=1}^{N} \big), \sigma \big( \{\hat{y}^{\mathcal{T},\mathrm{coarse}}_j\}_{j=1}^{N} \big) \Big) \\
& + D_{\mathrm{KL}} \Big( \sigma \big ( \{\hat{y}^{\mathcal{T},\mathrm{avg}}_j\}_{j=1}^{N} \big), \sigma \big( \{\hat{y}^{\mathcal{T},\mathrm{fine}}_j\}_{j=1}^{N} \big) \Big) \bigg) .
\end{split}
\end{equation}

The overall learning objective can be summarized as:
\begin{equation} \label{eq6}
\min \limits_{T,C_{\mathrm{coarse}},C_{\mathrm{fine}}} \ \mathcal{L}_{\mathrm{coarse}} + \mathcal{L}_{\mathrm{fine}} + \lambda \mathcal{L}_{\mathrm{distill}} ,
\end{equation}
where $\lambda$ is the trade-off parameter balancing between highlight prediction and knowledge distillation losses. 


\textbf{Inference scheme.} During inference, given a segment $s$ from a target category video, we first extend it into a set with other $N-1$ context segments adjacent to $s$ in the same video, and the set is denoted as $x_{\mathcal{T}} = \{s^{\mathcal{T}}_j\}_{j=1}^{N}$. The selection of these context segments follows the scheme depicted in the inference part of Sec.~\ref{sec3_2}. The highlight score of each segment in $x_{\mathcal{T}}$ is respectively inferred by the coarse-grained and fine-grained learner, which derives the highlight prediction $\hat{y}_{\mathcal{T},\mathrm{coarse}} = \{\hat{y}^{\mathcal{T},\mathrm{coarse}}_j\}_{j=1}^{N}$ and $\hat{y}_{\mathcal{T},\mathrm{fine}} = \{\hat{y}^{\mathcal{T},\mathrm{fine}}_j\}_{j=1}^{N}$. These two kinds of predictions are further averaged to produce $\hat{y}_{\mathcal{T},\mathrm{avg}} = \{\hat{y}^{\mathcal{T},\mathrm{avg}}_j\}_{j=1}^{N}$. Finally, we pick out $\hat{y}^{\mathcal{T},\mathrm{avg}}_{\mathrm{id}(s)}$ ($\mathrm{id}(s)$ denotes the index of $s$ in $x_{\mathcal{T}}$) as the highlight score of segment $s$. 


\begin{table*}[t]
\begin{spacing}{1.0}
\centering
\small
\caption{Highlight detection results (mAP) of weakly-supervised and supervised methods on the YouTube Highlights dataset.} \label{youtube_highlights}
\setlength{\tabcolsep}{2.1mm}
\begin{tabular}{c|ccc|cccc}
    \toprule[1.0pt]
    \multirow{2}{*}{Category} & \multicolumn{3}{|c|}{Weakly-supervised Methods} & \multicolumn{4}{c}{Supervised Methods} \\
    \cline{2-8}
    & RRAE~\cite{robust_autoencoder} & LIM-s~\cite{learn_from_duration} & MINI-Net~\cite{mini-net} & Video2GIF~\cite{video2gif} & LSVM~\cite{youtube_highlights} & SL-module (w/o $T$) & SL-module \\
    \hline
    dog & 0.49 & 0.579 & 0.537 & 0.308 & 0.60 & 0.690 & \textbf{0.708} \\
    gymnastics & 0.35 & 0.417 & 0.528 & 0.335 & 0.41 & 0.506 & \textbf{0.532} \\
    parkour & 0.50 & 0.670 & 0.689 & 0.540 & 0.61 & 0.690 & \textbf{0.772} \\
    skating & 0.25 & 0.578 & 0.709 & 0.554 & 0.62 & 0.687 & \textbf{0.725} \\
    skiing & 0.22 & 0.486 & 0.583 & 0.328 & 0.36 & 0.636 & \textbf{0.661} \\
    surfing & 0.49 & 0.651 & 0.638 & 0.541 & 0.61 & 0.695 & \textbf{0.762} \\
    \hline
    Average & 0.383 & 0.564 & 0.614 & 0.464 & 0.536 & 0.651 & \textbf{0.693} \\
    \bottomrule[1.0pt]
\end{tabular}
\end{spacing}
\end{table*}


\begin{table*}[t]
\begin{spacing}{1.1}
\centering
\scriptsize
\caption{Highlight detection results (top-5 mAP score) of weakly-supervised and supervised methods on the TVSum dataset.} \label{tvsum}
\setlength{\tabcolsep}{1.58mm}
\begin{tabular}{c|ccccc|cccccc}
    \toprule[1.0pt]
    \multirow{2}{*}{Category} & \multicolumn{5}{|c|}{Weakly-supervised Methods} & \multicolumn{6}{c}{Supervised Methods} \\
    \cline{2-12}
    & SG~\cite{adversarial_lstm} & DSN~\cite{summarize_web_video} & VESD~\cite{variational} & LIM-s~\cite{learn_from_duration} & MINI-Net~\cite{mini-net} & KVS~\cite{category-specific} & DPP~\cite{diverse_sequential} & vsLSTM~\cite{vslstm} & SM~\cite{submodular} & SL-module (w/o $T$) & SL-module \\
    \hline
    VT & 0.423 & 0.373 & 0.447 & 0.559 & 0.803 & 0.353 & 0.399 & 0.411 & 0.415 & 0.837 & \textbf{0.865} \\
    VU & 0.472 & 0.441 & 0.493 & 0.429 & 0.653 & 0.441 & 0.453 & 0.462 & 0.467 & 0.663 & \textbf{0.687} \\
    GA & 0.475 & 0.428 & 0.496 & 0.612 & \textbf{0.754} & 0.402 & 0.457 & 0.463 & 0.469 & 0.724 & 0.749 \\
    MS & 0.489 & 0.436 & 0.503 & 0.540 & 0.813 & 0.417 & 0.462 & 0.477 & 0.478 & 0.851 & \textbf{0.862} \\
    PK & 0.456 & 0.411 & 0.478 & 0.604 & 0.780 & 0.382 & 0.437 & 0.448 & 0.445 & 0.767 & \textbf{0.790} \\
    PR & 0.473 & 0.417 & 0.485 & 0.475 & 0.545 & 0.403 & 0.446 & 0.461 & 0.458 & 0.594 & \textbf{0.632} \\
    FM & 0.464 & 0.412 & 0.487 & 0.432 & 0.559 & 0.397 & 0.442 & 0.452 & 0.451 & 0.580 & \textbf{0.589} \\
    BK & 0.417 & 0.368 & 0.441 & 0.663 & 0.717 & 0.342 & 0.395 & 0.406 & 0.407 & 0.708 & \textbf{0.726} \\
    BT & 0.483 & 0.435 & 0.492 & 0.691 & 0.769 & 0.419 & 0.464 & 0.471 & 0.473 & 0.779 & \textbf{0.789} \\
    DS & 0.466 & 0.416 & 0.488 & 0.626 & 0.591 & 0.394 & 0.449 & 0.455 & 0.453 & 0.612 & \textbf{0.640} \\
    \hline
    Average & 0.462 & 0.424 & 0.481 & 0.563 & 0.698 & 0.398 & 0.447 & 0.451 & 0.461 & 0.712 & \textbf{0.733} \\
    \bottomrule[1.0pt]
\end{tabular}
\end{spacing}
\end{table*}


\begin{table}[t]
\begin{spacing}{1.1}
\centering
\scriptsize
\caption{Highlight detection results (mAP) of weakly-supervised and supervised methods on the ActivityNet dataset. 
} \label{activitynet}
\setlength{\tabcolsep}{0.26mm}
\begin{tabular}{c|cc|ccc}
    \toprule[1.0pt]
    \multirow{2}{*}{Category} & \multicolumn{2}{|c|}{Weakly-supervised} & \multicolumn{3}{c}{Supervised} \\
    \cline{2-6}
    & LIM-s~\cite{learn_from_duration} & MINI-Net~\cite{mini-net} & LSVM~\cite{youtube_highlights} & SL-module (w/o $T$) & SL-module \\
    \hline
    eat\&drink & 0.638 & 0.702 & 0.670 & 0.716 & \textbf{0.736} \\
    personal care & 0.663 & 0.689 & 0.657 & 0.725 & \textbf{0.744} \\
    household & 0.621 & 0.745 & 0.707 & 0.763 & \textbf{0.787} \\
    sport & 0.710 & 0.794 & 0.769 & 0.835 & \textbf{0.849} \\
    social & 0.743 & 0.760 & 0.740 & 0.758 & \textbf{0.779} \\
    \hline
    Average & 0.675 & 0.738 & 0.709 & 0.759 & \textbf{0.779} \\
    \bottomrule[1.0pt]
\end{tabular}
\end{spacing}
\end{table}


\section{Experiments} \label{sec4}

In this section, we compare the proposed \emph{SL-module} and the \emph{DL-VHD} method with existing video highlight detection approaches under the category-specific and cross-category setting, respectively.


\subsection{Experimental Setup} \label{sec4_1}

\textbf{Model details.} Following \cite{video2gif,learn_from_duration}, a C3D model~\cite{c3d} pre-trained on the UCF101 dataset~\cite{ucf101} serves as the backbone for feature extraction, and its parameters are fixed during training. The Transformer encoder is constructed with 5 layers of self-attention and feed forward block, and each multi-head self-attention module is equipped with 8 attention heads. The scoring model $C$, coarse-grained learner $C_{\mathrm{coarse}}$ and fine-grained learner $C_{\mathrm{fine}}$ are all instantiated as a multi-layer perceptron with architecture FC(4096,1024) $\rightarrow$ ReLU $\rightarrow$ FC(1024,256) $\rightarrow$ ReLU $\rightarrow$ FC(256,1), where FC is short for fully-connected layer. 


\textbf{Training details.} In all experiments, an SGD optimizer (initial learning rate: 0.001, momentum: 0.9, weight decay: $5 \times 10^{-4}$) is employed to train the model for 50 epochs, and the learning rate is multiplied by 0.1 every 20 epochs. For each video segment, 16 frames are sampled from it with the same interval. Without otherwise specified, the set size $N$ is set as 20, and the trade-off parameter $\lambda$ is set as $1.0$ (parameter sensitivity is analyzed in Sec.~\ref{sec5_2}). We use an NVIDIA Tesla V100 GPU for training. 
Our method is implemented with the PyTorch \cite{pytorch} deep learning framework, and the source code will be released for reproducibility. 


\textbf{Performance comparison.} Under the category-specific setting, six supervised video highlight detection (or video summarization) methods, \emph{i.e.} Video2GIF~\cite{video2gif}, LSVM~\cite{youtube_highlights}, KVS~\cite{category-specific}, DPP~\cite{diverse_sequential}, vsLSTM~\cite{vslstm} and SM~\cite{submodular}, and six weakly-supervised approaches, \emph{i.e.} RRAE~\cite{robust_autoencoder}, SG~\cite{adversarial_lstm}, DSN~\cite{summarize_web_video}, VESD~\cite{variational}, LIM-s~\cite{learn_from_duration} and MINI-Net~\cite{mini-net}, are introduced for comparison. For the cross-category setting, the SL-module trained on the source/target video category serves as the lower/upper bound of model performance. For the sake of fair comparison, five UDA algorithms, \emph{i.e.} DAN~\cite{dan}, DeepCORAL~\cite{deepcoral}, RevGrad~\cite{revgrad}, MCD~\cite{max_discrepancy} and AFN~\cite{larger_norm}, are combined with SL-module to compare with the proposed DL-VHD method, and the detailed combination schemes are provided in the supplementary material. 


\subsection{Category-specific Video Highlight Detection} \label{sec4_2}

When highlight annotations are available on the video category intended to be used, SL-module can be individually applied to train a highlight detector under the category-specific setting. We compare it with existing video highlight detection and video summarization methods in this section.

\textbf{Datasets.} 
\emph{YouTube Highlights}~\cite{youtube_highlights} is composed of six video categories, \emph{i.e.} dog, gymnastics, parkour, skating, skiing and surfing, and each category has approximately 100 videos. Segment-level annotations are provided to indicate whether a segment is a highlight moment or not. We follow the standard training-test split~\cite{youtube_highlights} for model evaluation. 

\emph{TVSum}~\cite{tvsum} is a video summarization dataset consisting of 10 categories of video events with 5 videos in each category, and frame-level importance score is provided in this dataset. Following previous works~\cite{learn_from_duration,mini-net}, we average the frame-level importance scores to achieve the segment-level highlight scores. 
For each video category, we select the two longest videos (about 10 minutes in total) for training and the rest three ones for test. 

\emph{ActivityNet}~\cite{activitynet} is a large-scale database for human activity classification and detection. We employ the data of the temporal action localization track for highlight detection. Specifically, we split the video samples to five categories, \emph{i.e.} eat\&drink, personal care, household, sport and social, according to the first-level action label. The temporal Intersection over Union (tIoU) between a video segment and a ground-truth event of a specific category is used as the segment's highlight label for this video category. Totally, we utilize 2520 videos for training and 1260 videos for test, and the detailed dataset statistics for all video categories are provided in the supplementary material. 


\textbf{Results on YouTube Highlights.} In Tab.~\ref{youtube_highlights}, we compare our method with existing approaches on six video categories of YouTube Highlights. It can be observed that the proposed SL-module outperforms previous pair-learning-based algorithms, \emph{i.e.} LIM-s, MINI-Net, Video2GIF and LSVM, on all six categories, and superior average mAP is still obtained when the Transformer encoder $T$ is removed from our model. This phenomenon illustrates the superiority of set-based learning over pair-based methods, in which the broader contextual information within a segment set enables more precise highlight prediction of each video segment. 


\textbf{Results on TVSum.} Tab.~\ref{tvsum} reports the performance of various video highlight detection and video summarization approaches on TVSum. On nine of ten video categories, the proposed SL-module achieves the best performance, and, when removing the Transformer encoder, it still outperforms the state-of-the-art MINI-Net on seven of ten categories. These results verify the effectiveness of set-based learning under the circumstances with limited training data, \emph{i.e.} only two videos per category for training. 


\textbf{Results on ActivityNet.} In Tab.~\ref{activitynet}, we evaluate the performance of three existing methods and two configurations of the proposed model. Since the experiments on ActivityNet dataset were not commonly included in previous works, we examine these works by the released source code (for MINI-Net and LSVM) or our re-implementation (for LIM-s). The experimental results on this large-scale dataset further verify the superiority of the proposed set-learning method (\emph{i.e.} obtaining the highest test mAP on all five video categories) when the training data is abundant. 


\begin{table}[t]
\begin{spacing}{1.1}
\centering
\scriptsize
\caption{Cross-category highlight detection results (mAP) on the YouTube Highlights dataset. (source video category: surfing; the \underline{underlined} result surpasses the target-oracle.)} \label{surfing_transfer}
\setlength{\tabcolsep}{0.65mm}
\begin{tabular}{c|ccccc}
    \toprule[1.0pt]
    Methods & $\rightarrow$dog & $\rightarrow$gymnastics & $\rightarrow$parkour & $\rightarrow$skating & $\rightarrow$skiing \\
    \hline
    \hline
    Source-only & 0.485 & 0.505 & 0.547 & 0.568 & 0.545 \\
    \hline
    DAN~\cite{dan} & \textbf{0.652} & 0.487 & 0.713 & 0.638 & 0.611 \\
    DeepCORAL~\cite{deepcoral} & 0.634 & 0.513 & 0.732 & 0.659 & 0.620 \\
    RevGrad~\cite{revgrad} & 0.628 & 0.493 & 0.654 & 0.640 & 0.597 \\
    MCD~\cite{max_discrepancy} & 0.567 & 0.529 & 0.499 & 0.642 & 0.654 \\
    AFN~\cite{larger_norm} & 0.625 & 0.517 & 0.575 & 0.653 & 0.626 \\
    \hline
    DL-VHD ($\mathcal{L}_{\mathrm{coarse}}$ only) & 0.574 & 0.498 & 0.704 & 0.635 & 0.631 \\
    DL-VHD ($\mathcal{L}_{\mathrm{fine}}$ only) & 0.485 & 0.505 & 0.547 & 0.568 & 0.545 \\
    DL-VHD (w/o $\mathcal{L}_{\mathrm{distill}}$) & 0.630 & 0.529 & 0.718 & 0.683 & 0.658 \\
    DL-VHD (full model) & 0.649 & \textbf{\underline{0.540}} & \textbf{0.748} & \textbf{0.713} & \textbf{\underline{0.686}} \\
    \hline 
    Target-oracle & 0.708 & 0.532 & 0.772 & 0.725 & 0.661 \\
    \bottomrule[1.0pt]
\end{tabular}
\end{spacing}
\end{table}


\subsection{Cross-category Video Highlight Detection} \label{sec4_3}

Under the cross-category highlight detection setting, we evaluate the effectiveness of DL-VHD and various UDA algorithms on transferring the highlight knowledge from source video category to the target one. In all experiments, the videos of source category possess segment-level annotations, while the videos of target category are unannotated.

\textbf{Tasks.} 
\emph{YouTube Highlights} consists of six video categories, and we employ \emph{surfing} as the source category and evaluate each of the cases that one of the other five categories serves as the target one. Also, we consider a more difficult setting where \emph{dog} is used as the source category (\emph{i.e.} adapting from dog activities to human ones), and the results of this setting are in the supplementary material. 

\emph{ActivityNet} contains five categories of human activities, and we utilize \emph{sport} as the source category and aim at transferring the knowledge of sport highlights to other four video categories. The adaptation towards each target video category is separately examined.


\begin{table}[t]
\begin{spacing}{1.1}
\centering
\scriptsize
\caption{Cross-category highlight detection results (mAP) on the ActivityNet dataset. (source video category: sport; the \underline{underlined} result surpasses the target-oracle.)} \label{sport_transfer}
\setlength{\tabcolsep}{0.76mm}
\begin{tabular}{c|cccc}
    \toprule[1.0pt]
    Methods & $\rightarrow$eat\&drink & $\rightarrow$personal care & $\rightarrow$household & $\rightarrow$social \\
    \hline
    \hline
    Source-only & 0.674 & 0.667 & 0.707 & 0.722 \\
    \hline
    DAN~\cite{dan} & 0.656 & 0.678 & 0.694 & 0.735 \\
    DeepCORAL~\cite{deepcoral} & 0.708 & 0.705 & 0.765 & 0.744 \\
    RevGrad~\cite{revgrad} & 0.687 & 0.701 & 0.722 & 0.731 \\
    MCD~\cite{max_discrepancy} & 0.712 & 0.713 & 0.761 & 0.756 \\
    AFN~\cite{larger_norm} & 0.718 & 0.704 & 0.750 & 0.749 \\
    \hline
    DL-VHD ($\mathcal{L}_{\mathrm{coarse}}$ only) & 0.689 & 0.694 & 0.742 & 0.741 \\
    DL-VHD ($\mathcal{L}_{\mathrm{fine}}$ only) & 0.674 & 0.667 & 0.707 & 0.722 \\
    DL-VHD (w/o $\mathcal{L}_{\mathrm{distill}}$) & 0.713 & 0.715 & 0.778 & 0.754 \\
    DL-VHD (full model) & \textbf{0.730} & \textbf{0.728} & \textbf{\underline{0.793}} & \textbf{0.766} \\
    \hline 
    Target-oracle & 0.736 & 0.744 & 0.787 & 0.779 \\
    \bottomrule[1.0pt]
\end{tabular}
\end{spacing}
\end{table}


\textbf{Cross-category results on YouTube Highlights.} Tab.~\ref{surfing_transfer} reports the performance of various approaches on five cross-category highlight detection tasks, in which \emph{surfing} serves as the source category. Source-only (target-oracle) method represents the SL-module trained on the source (target) video category in a supervised fashion, where an obvious performance gap exists between them.
We can observe that the full model of DL-VHD surpasses five existing UDA algorithms on four of five tasks, and it surprisingly outperforms the target-oracle model on two tasks, \emph{i.e.} surfing $\rightarrow$ gymnastics and surfing $\rightarrow$ skiing. Such results illustrate that cross-category video highlight detection cannot be easily deemed as a variant of UDA problem, and more dedicated techniques (\emph{e.g.} the proposed dual-learner and knowledge distillation schemes) can better discover the transferrable highlight patterns across different video categories. 


\textbf{Cross-category results on ActivityNet.} In Tab.~\ref{sport_transfer}, we compare the proposed DL-VHD model with five UDA methods on the cross-category highlight detection tasks of ActivityNet, and \emph{sport} is utilized as the source category in all these tasks. The full model of DL-VHD achieves higher mAP than the UDA algorithms on all four tasks, and it even outperforms the target-oracle model on the sport $\rightarrow$ household task. 
These empirical results verify that the DL-VHD model succeeds in capturing the human-related action patterns on the target video category under the guidance of labeled source videos and unlabeled target videos. 


\section{Analysis}  \label{sec5}

In this section, we conduct more in-depth analysis of our approach to evaluate the effectiveness of major model components both quantitatively and qualitatively. 


\begin{figure*}[t]
    \centering
    \includegraphics[width=0.97\textwidth]{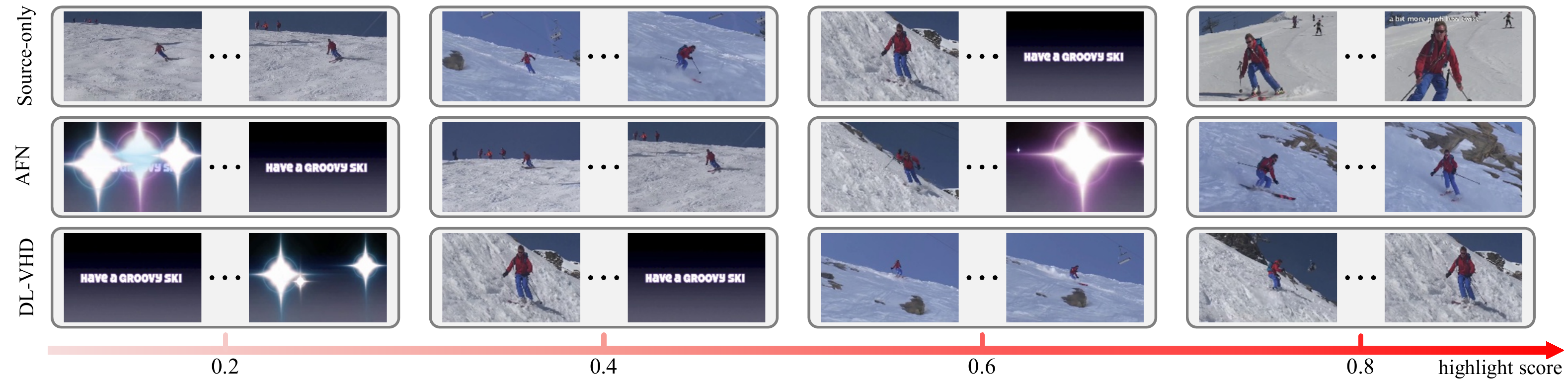}
    \caption{Highlight predictions of three methods on the \emph{surfing} $\rightarrow$ \emph{skiing} task. (Each video segment is denoted by its first and last frames.)} 
    \label{fig_visualization}
\vspace{-3mm}
\end{figure*}


\begin{figure}[t]
    \centering
    \includegraphics[width=0.47\textwidth]{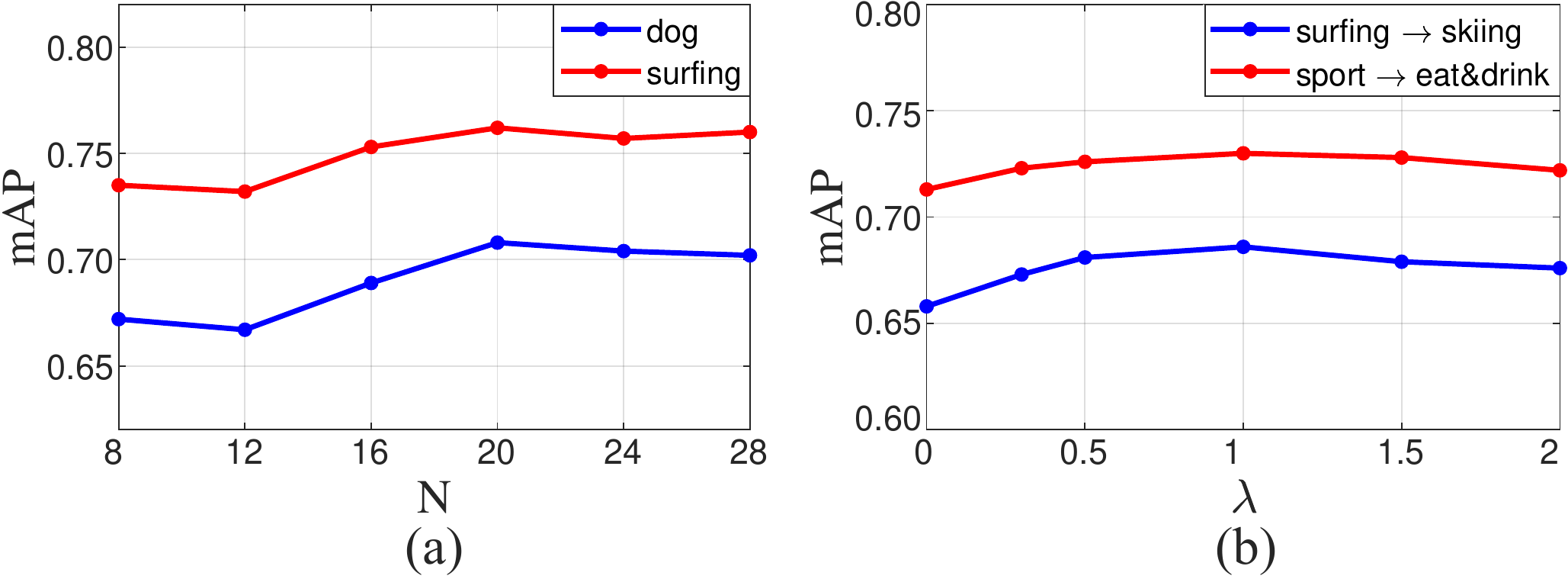}
    \caption{Sensitivity analysis of set size $N$ (left) and trade-off parameter $\lambda$ (right).} 
    \label{fig_sensitivity}
\vspace{-3mm}
\end{figure}


\subsection{Ablation Study} \label{sec5_1}

\textbf{Effect of Transformer encoder.} In all three video highlight detection datasets, we compare the performance of the SL-module with and without Transformer encoder $T$, as shown in Tabs.~\ref{youtube_highlights}, \ref{tvsum} and \ref{activitynet}. It can be observed that, after applying the Transformer encoder, the proposed set-based learning method obtains a clear performance gain on all tasks, which demonstrates the importance of interrelationship modeling when learning from a set of video segments. 


\textbf{Effect of dual learners and knowledge distillation.} In Tabs.~\ref{surfing_transfer} and \ref{sport_transfer}, we investigate the main components of DL-VHD through three additional model configurations: (1)~$\mathcal{L}_{\mathrm{coarse}}$ only: only the coarse-grained learner is utilized to predict the highlight extent of a segment with respect to the target video category; (2) $\mathcal{L}_{\mathrm{fine}}$ only: only the fine-grained learner is employed for highlight prediction on target category (this configuration is equivalent to the source-only baseline); (3) w/o $\mathcal{L}_{\mathrm{distill}}$: both the coarse- and fine-grained learners are trained, while their knowledge is not integrated by the knowledge distillation loss. When the two learners are individually applied, the coarse-grained learner outperforms the fine-grained one, which, we think, is because the supervision for coarse-grained learner is more relevant to the highlight patterns on target video category than the supervision applied to fine-grained learner. In the full model, the knowledge distillation scheme is able to further promote model's performance upon configuration (3) by integrating the knowledge of two learners. 


\subsection{Sensitivity Analysis} \label{sec5_2}

\textbf{Sensitivity of set size $N$.} In this experiment, we analyze the sensitivity of the proposed SL-module to the set size. Fig.~\ref{fig_sensitivity}(a) shows the model performance on two highlight detection tasks under different set sizes. It can be observed that our set-based learning method can achieve stable performance gain when the size of each segment set is large enough, \emph{i.e.} $N \geqslant 16$. 


\textbf{Sensitivity of trade-off parameter $\lambda$.} In this part, we discuss the selection of trade-off parameter $\lambda$ which balances between highlight prediction and knowledge distillation objectives. In Fig.~\ref{fig_sensitivity}(b), we plot the performance of DL-VHD on two cross-category highlight detection tasks using various $\lambda$ values. The highest mAP on target video category is gained when the value of $\lambda$ is around $1.0$, which indicates that the appropriate balance between two distinct optimization objectives is attained under such condition. 


\subsection{Visualization} \label{sec5_3}

For the cross-category highlight detection task \emph{surfing} $\rightarrow$ \emph{skiing}, Fig.~\ref{fig_visualization} visualizes the highlight prediction results of three methods, \emph{i.e.} source-only, AFN and DL-VHD, on a target category video.
For each method, we select the segments with the closest highlight score to the corresponding coordinate value (0.2, 0.4, 0.6 or 0.8), and each segment is represented by its first and last frames. The source-only model fails to capture the highlight patterns of skiing, and the AFN algorithm performs better but still overvalue a non-highlight segment by a score near 0.6. By comparison, DL-VHD assigns highlight scores to various video segments most appropriately. More visualization results on other tasks can be found in the supplementary material. 


\section{Conclusions and Future Work}  \label{sec6}

In this research, we novelly explore the cross-category video highlight detection problem with a Dual-Learner-based Video Highlight Detection (DL-VHD) framework. Under this framework, a Set-based Learning module (SL-module) is proposed to improve the commonly employed pair-based learning, and dual-learner and knowledge distillation schemes are further introduced for highlight knowledge transfer. The comprehensive experiments under both the category-specific and cross-category settings verify the exceeding performance of the proposed method.

Our future explorations will involve further improving the algorithm for cross-category highlight detection, applying the proposed approach to more sophisticated real-world applications and studying on the generalization capability of video highlight detection models. 


\section{Acknowledgement} \label{sec7}

This work was supported by National Science Foundation of China (U20B2072, 61976137). Authors appreciate the Student Innovation Center of SJTU and the ByteDance AI Lab for providing GPUs. Authors also thank Jie Zhou, Jiawen Li and Xuanyu Zhu for their valuable suggestions. 


\newpage
{\small
\bibliographystyle{ieee_fullname}
\bibliography{reference}
}


\begin{figure*}[t]
    \centering
    \includegraphics[width=0.97\textwidth]{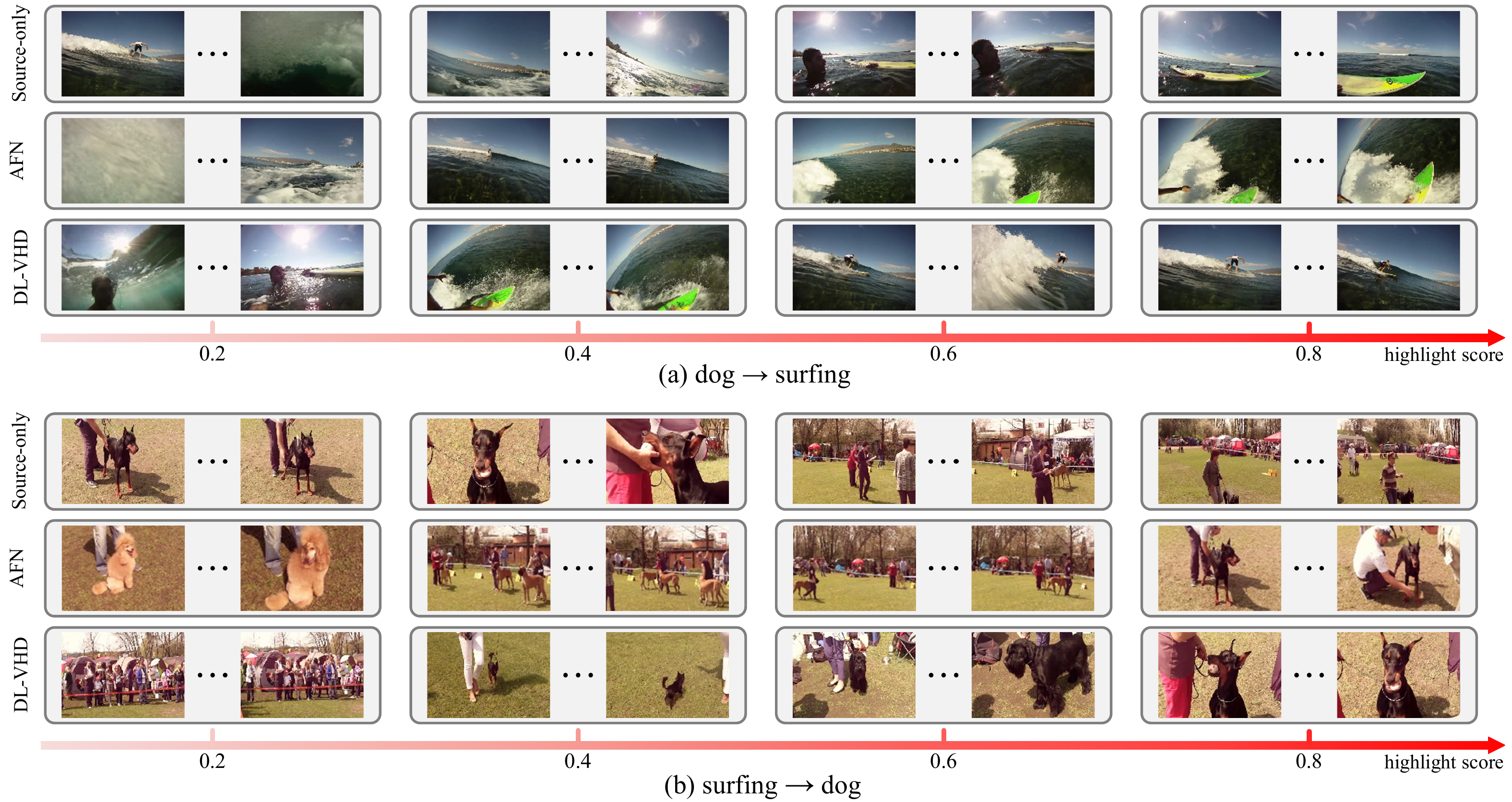}
    \caption{Highlight predictions of three methods on two cross-category highlight detection tasks, \emph{i.e.} \emph{dog} $\rightarrow$ \emph{surfing} and \emph{surfing} $\rightarrow$ \emph{dog}. (Each video segment is represented by its first and last frames.)} 
    \label{fig_supp_visualization}
    \vspace{-2mm}
\end{figure*}


\newpage
\begin{table}[t]
\begin{spacing}{1.1}
\centering
\small
\caption{Statistics of the ActivityNet dataset in our experiments.} \label{activitynet_statistics}
\setlength{\tabcolsep}{0.7mm}
\begin{tabular}{c|ccccc|c}
    \toprule[1.0pt]
    Split & eat\&drink & personal care & household & sport & social & Total \\
    \hline
    \hline
    Training & 140 & 186 & 458 & 1289 & 447 & 2520 \\
    \hline
    Test & 65 & 95 & 212 & 672 & 216 & 1260 \\
    \bottomrule[1.0pt]
\end{tabular}
\end{spacing}
\end{table}


\section{More Experimental Setups} \label{supp_sec1}

\textbf{Combining SL-module with UDA methods.} For the sake of fair comparison, we combine five Unsupervised Domain Adaptation (UDA) algorithms, \emph{i.e.} DAN~\cite{dan}, DeepCORAL~\cite{deepcoral}, RevGrad~\cite{revgrad}, MCD~\cite{max_discrepancy} and AFN~\cite{larger_norm}, with the proposed SL-module and compare these combinations with the DL-VHD method. DAN, DeepCORAL and AFN align the feature distributions of source and target domain by minimizing specific domain discrepancy metrics, and we exert these metric-induced alignment losses on the contextualized segment embeddings (\emph{i.e.} outputs of Transformer encoder) to narrow the distributional gap between source and target category video segments in the latent space. For RevGrad, we append a domain discriminator on the top of contextualized segment embeddings to conduct adversarial domain adaptation. For MCD, we train two scoring models on the source video category in a supervised way, and a minimax game is performed between Transformer encoder and two scoring models to derive more reliable highlight predictions on target video category. 


\textbf{Dataset statistics of ActivityNet.} In our experiments, we employ a subset of ActivityNet~\cite{activitynet} for model evaluation. The number of videos in the training and test split for each video category is shown in Tab.~\ref{activitynet_statistics}. 
Note that, each of these videos contains at least one highlight moment of the corresponding video category. 


\begin{table}[t]
\begin{spacing}{1.1}
\centering
\scriptsize
\caption{Cross-category highlight detection results (mAP) on the YouTube Highlights dataset. (source video category: dog; the \underline{underlined} result surpasses the target-oracle.)} \label{dog_transfer}
\setlength{\tabcolsep}{0.42mm}
\begin{tabular}{c|ccccc}
    \toprule[1.0pt]
    Methods & $\rightarrow$gymnastics & $\rightarrow$parkour & $\rightarrow$skating & $\rightarrow$skiing & $\rightarrow$surfing \\
    \hline
    \hline
    Source-only & 0.486 & 0.480 & 0.535 & 0.564 & 0.531 \\
    \hline
    DAN~\cite{dan} & 0.520 & 0.674 & 0.632 & 0.613 & 0.575 \\
    DeepCORAL~\cite{deepcoral} & 0.518 & 0.615 & 0.615 & 0.609 & 0.517 \\
    RevGrad~\cite{revgrad} & 0.514 & 0.630 & 0.629 & 0.618 & 0.587 \\
    MCD~\cite{max_discrepancy} & 0.479 & 0.587 & 0.658 & 0.614 & 0.625 \\
    AFN~\cite{larger_norm} & 0.498 & 0.594 & 0.607 & 0.620 & 0.589 \\
    \hline
    DL-VHD ($\mathcal{L}_{\mathrm{coarse}}$ only) & 0.489 & 0.495 & 0.571 & 0.608 & 0.559 \\
    DL-VHD ($\mathcal{L}_{\mathrm{fine}}$ only) & 0.486 & 0.480 & 0.535 & 0.564 & 0.531 \\
    DL-VHD (w/o $\mathcal{L}_{\mathrm{distill}}$) & 0.525 & 0.686 & 0.654 & 0.630 & 0.649 \\
    DL-VHD (full model) & \textbf{\underline{0.556}} & \textbf{0.734} & \textbf{0.692} & \textbf{0.653} & \textbf{0.676} \\
    \hline 
    Target-oracle & 0.532 & 0.772 & 0.725 & 0.661 & 0.762 \\
    \bottomrule[1.0pt]
\end{tabular}
\end{spacing}
\end{table}


\section{More Results of Cross-category Video \\ Highlight Detection} \label{supp_sec2}

In Tab.~\ref{dog_transfer}, we evaluate different methods on five cross-category video highlight detection tasks of YouTube Highlights~\cite{youtube_highlights}, in which \emph{dog} serves as the source video category. This setting is more difficult than the one employing \emph{surfing} as the source category, since it intends to transfer the highlight patterns of dog to human. Source-only (target-oracle) method denotes the SL-module trained on the source (target) video category in a supervised way. From the table, we can observe that the full model of DL-VHD outperforms five UDA approaches with a clear margin, and it surpasses the target-oracle model on the dog $\rightarrow$ gymnastics task. When the coarse-grained or fine-grained learner is individually applied (\emph{i.e.} the configuration $\mathcal{L}_\mathrm{coarse}$ only and $\mathcal{L}_\mathrm{fine}$ only), their performance is 
apparently lower than their combination (\emph{i.e.} the configuration w/o $\mathcal{L}_\mathrm{distill}$ and full model). After integrating the knowledge of two learners, the full model can derive more precise highlight predictions on target video category than the model without applying knowledge distillation.


\section{More Visualization Results} \label{supp_sec3}

Fig.~\ref{fig_supp_visualization} visualizes the highlight prediction results of three approaches on the target category video of two more difficult tasks, \emph{i.e.} \emph{dog} $\rightarrow$ \emph{surfing} and \emph{surfing} $\rightarrow$ \emph{dog}. Compared to source-only and AFN~\cite{larger_norm}, the proposed DL-VHD method can better acquire the concepts about the highlight moments on target video category, \emph{e.g.} the segments describing an athlete surfing on the wave or the ones containing dogs.


\end{document}